\pdfoutput=1
\documentclass[final,onefignum,onetabnum,hidelinks]{siamart251216}

\usepackage[utf8]{inputenc}
\usepackage{microtype,amssymb,bm,mathtools}
\usepackage{graphicx,booktabs,array,tabularx,adjustbox}
\usepackage[compatibility=false]{caption}
\usepackage{siunitx,placeins,url}
\DeclareUnicodeCharacter{03C7}{\ensuremath{\chi}}
\crefname{appendix}{Appendix}{Appendices}
\Crefname{appendix}{Appendix}{Appendices}
\newcommand{\vect}[1]{\mathbf{#1}}
\newcommand{\mat}[1]{\mathbf{#1}}
\newcommand{\IdMat}[1]{\mat{I}_{#1}}
\newcommand{\Rdim}[1]{\mathbb{R}^{#1}}
\newcommand{\wnorm}[2]{\left\lVert #1 \right\rVert_{#2^{-1}}^2}
\newcommand{\rv}[1]{\mathsf{#1}}
\newcommand{\IntVar}{\xi}
\newcommand{\E}{\mathbb{E}}
\newcommand{\dd}{\mathrm{d}}
\newcommand{\Normal}[2]{\mathcal{N}\!\left(#1,\,#2\right)}
\newcommand{\DKL}[2]{\mathcal{D}_{\mathrm{KL}}\!\left(#1 \| #2 \right)}
\newcommand{\DTV}[2]{\mathcal{D}_{\mathrm{TV}}\!\left(#1 \| #2 \right)}
\newcommand{\Entropy}[1]{H\!\left(#1\right)}
\newcommand{\param}{\theta}
\newcommand{\ctx}{u}
\newcommand{\obs}{y}
\newcommand{\parami}[1]{\param_{#1}}
\newcommand{\ctxi}[1]{\ctx_{#1}}
\newcommand{\Param}{\bm{\theta}}
\newcommand{\Ctx}{\vect{\ctx}}
\newcommand{\Obs}{\vect{\obs}}
\newcommand{\ParamSpace}{\Theta}
\newcommand{\CtxSpace}{\mathcal{U}}
\newcommand{\ObsSpace}{\mathcal{Y}}
\newcommand{\Obsdim}{d_{\obs}}
\newcommand{\Ctxi}[1]{\Ctx_{#1}}
\newcommand{\Obsi}[1]{\Obs_{#1}}
\newcommand{\Ctxn}{\Ctx_{1:n}}
\newcommand{\Obsn}{\Obs_{1:n}}
\newcommand{\ObsNoise}{\vect{\varepsilon}}
\newcommand{\rvObsNoiseCov}{\mat{R}}
\newcommand{\CovPrior}{\mat{B}}
\newcommand{\rvParam}{{\bm{\Theta}}}
\newcommand{\rvCtx}{\rv{U}}
\newcommand{\rvObs}{\rv{Y}}
\newcommand{\rvCtxi}[1]{\rvCtx_{#1}}
\newcommand{\rvObsi}[1]{\rvObs_{#1}}
\newcommand{\rvCtxn}{\rvCtx_{1:n}}
\newcommand{\rvObsn}{\rvObs_{1:n}}
\newcommand{\CtxSingle}{\Ctxi{s}}
\newcommand{\ObsSingle}{\Obsi{s}}
\newcommand{\ModelObs}{\mathcal{G}}
\newcommand{\ModelObsCall}[2]{\ModelObs\!\left(#1,\,#2\right)}
\newcommand{\Paramb}{\Param_b}
\newcommand{\eqc}{\overset{c}{=}}
\newcommand{\VarParam}{\phi}
\newcommand{\qvar}{q_{\VarParam}}
\newcommand{\qvarof}[1]{q_{\VarParam}(#1)}
\newcommand{\NELBO}{\mathcal{L}_{\mathrm{ELBO}}}
\newcommand{\NELBOof}[3]{\NELBO(#1; #2, #3)}
\newcommand{\FlowMapName}{\mathcal{T}}
\newcommand{\rvBase}{\rv{Z}}
\newcommand{\BaseLatent}{q_{\rvBase}}
\newcommand{\FlowMapVI}{\FlowMapName_{\VarParam}}
\newcommand{\ParamTrue}{\Param^\mathrm{Truth}}
\newcommand{\RegNELBO}{\mathcal{L}_{\mathrm{ELBO},r}}
\newcommand{\RegNELBOCall}[3]{\RegNELBO\!\left(#1; #2, #3\right)}
\newcommand{\RegTerm}[1]{\mathcal{R}_{#1}}
\newcommand{\Renyi}{\mathcal{D}_{\mathrm{Renyi}}}
\newcommand{\RenyiCall}[3]{\Renyi^{#1}\!\left(#2 \,\|\, #3\right)}
\newcommand{\RegWeight}{\lambda}
\newcommand{\RenyiParam}{\rho}
\newcommand{\EMAParam}{\beta}
\newcommand{\Stab}{\epsilon}
\newcommand{\GradRatio}{r}
\hypersetup{hidelinks,unicode,pdfauthor={Exauc\'e Luweh Adjim Ngarti}}
\title{R\'enyi-Regularized Flow-Based Variational Inference for Bayesian Calibration}
\newcommand{\paperauthors}{%
  Exauc\'e Luweh Adjim Ngarti\textsuperscript{*},
  Elise Arnaud, L\'eo Nicoletti\\[0.35em]
  Victor Trappler, Arthur Vidard%
}

\newcommand{\paperaffiliations}{%
  \begingroup
  \centering
  \textsuperscript{1} Univ. Grenoble Alpes, Inria, CNRS,
  LJK, 38000 Grenoble, France.\par
  \textsuperscript{2} Bull,
Centre of Excellence in Performance Programming (CEPP),
France.\par
  \textsuperscript{3} Department GMI, Institut Henri Fayol,
  Mines Saint-\'Etienne, Univ Clermont Auvergne,
  CNRS, UMR 6158 LIMOS,
  F-42023 Saint-\'Etienne, France.\par
  \endgroup
}
\author{\paperauthors\thanks{%
Corresponding author: Exauc\'e Luweh Adjim Ngarti.
Email: \href{mailto:exaucengarti@gmail.com}
{\nolinkurl{exaucengarti@gmail.com}}.}}
\headers{}{}

\makeatletter
\let\paper@thanks\thanks
\renewcommand{\thanks}[1]{\begingroup\let\@makefnmark\relax\paper@thanks{#1}\endgroup}
\makeatother

\begin{document}
\maketitle
\paperaffiliations
\vspace{1em}

\begin{abstract}
Variational inference with normalizing flows provides a flexible posterior approximation for Bayesian model calibration, but struggles when the posterior has several modes separated by regions of low probability mass. We study this behavior in a two-parameter twin experiment and regularize the variational objective. Under Gaussian observation errors and a Gaussian prior, the corresponding minimization objective is the expected Tikhonov calibration objective minus the differential entropy of the approximating distribution. However, the expected Tikhonov term favors concentration near one of the modes, while entropy maximization does not guarantee multimodality.

We introduce a regularization term based on R\'enyi divergence that encourages the approximate posterior to cover the prior support. In the connected bimodal case, a flow approximation represents both modes; in the case with well-separated modes, the flow and Gaussian mixture frequently miss one mode. The introduced regularization reduces the near-collapse fraction from 50\% to 25\%. While the regularization improves mode coverage in this experiment, it changes the Bayesian variational objective, introduces a trade-off in approximation accuracy, and does not provide a general solution to mode collapse.
\end{abstract}

\section{Introduction}\label{sec:introduction}

Bayesian calibration characterizes uncertainties in numerical model parameters through posterior distributions. Markov chain Monte Carlo methods, including Metropolis--Hastings, Gibbs sampling, and Hamiltonian Monte Carlo \cite{hastings1970monte,geman1984stochastic,brooks2011handbook,Neal2011}, provide samples from the posterior. However, their computational cost can be prohibitive when likelihood evaluations are expensive, as is often the case for numerical simulators.
Variational inference offers an optimization-based approximation \cite{jordan1999introduction,hoffman2013stochastic,blei2017variational} using a parameterized family of distributions, and normalizing flows provide flexible distribution models with tractable sampling and log-density evaluation \cite{rezende2015variational,dinh2016density}.
Alternative variational objectives include $\chi$-divergences \cite{dieng2017variational}, $\alpha$-divergences \cite{li_renyi_2016,hernandez-lobato_black-box_2016,daudel_alpha-divergence_2023}, robust and tail-adaptive $f$-divergences \cite{futami_variational_2018,wang_variational_2019}, and Tsallis divergence \cite{wang_variational_2021}. In this work, we consider the standard ELBO derived from the reverse Kullback--Leibler divergence.

However, ELBO-based variational inference struggles when the posterior exhibits multiple modes separated by regions of low probability mass.
We augment the negative ELBO with a R\'enyi divergence from the prior to the variational approximation to encourage the approximate posterior to cover the prior support.
We test the regularization in a two-parameter calibration problem with two benchmark cases: a curved bimodal distribution with no region of near-zero probability mass separating the modes, and a bimodal distribution with modes separated by a region of near-zero probability mass.
Variational inference with normalizing flows yields a good approximation in the connected-posterior case but struggles in the case with well-separated modes. The regularization reduces the near-collapse fraction from 50\% to 25\%.

We first recall the Bayesian calibration framework in \cref{sec:bayesian_calibration}, link the ELBO objective to classical Tikhonov regularization, and explain how this link accounts for the observed mode collapse for posteriors with modes separated by regions of low probability mass. Then, in \cref{sec:renyi_regularization}, we introduce the R\'enyi regularization and present numerical experiments on the controlled benchmark in \cref{sec:numerical_experiments}.

\section{Background}\label{sec:background}

\subsection{Bayesian model calibration}
\label{sec:bayesian_calibration}

Let $\ModelObs:\ParamSpace\times\CtxSpace\to\ObsSpace$ denote the numerical model, $\Param \in \ParamSpace$ the model parameter to be calibrated, and $\Ctx \in \CtxSpace$ the context variable specified for each model run. Calibration is performed using observations and their corresponding contexts $(\Obsn,\Ctxn)$. Assuming conditionally independent observations with Gaussian errors and a Gaussian prior on the calibration parameter independent of the context, we have
\begin{equation}\label{eq:likelihood_model}
\rvObsi{i}\mid\Param,\Ctxi{i}\sim\Normal{\ModelObsCall{\Param}{\Ctxi{i}}}{\rvObsNoiseCov},
\qquad \rvParam\sim\Normal{\Paramb}{\CovPrior},
\end{equation}
where $\Paramb \in \ParamSpace$ is the prior mean, and $\CovPrior$ and $\rvObsNoiseCov$ are positive-definite covariance matrices. The posterior density is
\begin{equation}\label{eq:bayesposterior}
\pi(\Param; \Obsn, \Ctxn):=p_{\rvParam\mid\rvObsn,\rvCtxn}(\Param\mid\Obsn,\Ctxn)
\propto p_{\rvParam}(\Param)\prod_{i=1}^{n}p_{\rvObsi{i}\mid\rvParam,\rvCtxi{i}}(\Obsi{i}\mid\Param,\Ctxi{i}).
\end{equation}

\subsection{Variational inference with normalizing flows}

Variational inference approximates the posterior in \cref{eq:bayesposterior} with a parameterized distribution $\qvar$ by minimizing the Kullback--Leibler divergence, equivalently minimizing the negative ELBO,
\begin{equation}\label{eq:nelbo_flow_vi}
\NELBOof{\VarParam}{\Obsn}{\Ctxn}
=\E_{\qvar}\!\left[\log \qvarof{\rvParam}
-
\log p_{\rvParam}\!\left(\rvParam\right)-\sum_{i=1}^{n} \log p_{\rvObsi{i} \mid \rvParam,\rvCtxi{i}}
\!\left(
\Obsi{i} \mid \rvParam,\Ctxi{i}
\right)\right].
\end{equation}
Here, $\qvar$ is modeled by a normalizing flow, $\qvar=(\FlowMapVI)_\sharp\BaseLatent$, with $\FlowMapVI$ a differentiable, invertible map that transforms a base distribution $\BaseLatent$ into a more expressive distribution.

\subsection{Connection with Tikhonov regularization}

For a positive-definite matrix $\CovPrior$, define the weighted quadratic form $\wnorm{v}{\CovPrior}:=v^\top \CovPrior^{-1}v$. The Gaussian calibration assumptions lead to the Tikhonov loss \cite{tikhonov1965use}
\begin{equation}\label{eq:tikhonov}
J(\Param)=\frac12\sum_{i=1}^{n}\wnorm{\Obsi{i}-\ModelObsCall{\Param}{\Ctxi{i}}}{\rvObsNoiseCov}
+\frac12\wnorm{\Param-\Paramb}{\CovPrior}.
\end{equation}
For the flow distribution, let $\Entropy{\qvar}=-\E_{\qvar}[\log\qvarof{\rvParam}]$ denote its differential entropy. Substituting the Gaussian likelihood and prior into \cref{eq:nelbo_flow_vi} gives
\begin{equation}\label{eq:nelbo_tikhonov}
\NELBOof{\VarParam}{\Obsn}{\Ctxn}\eqc\E_{\qvar}[J(\rvParam)]-\Entropy{\qvar},
\end{equation}
where $\eqc$ denotes equality up to a constant independent of $\VarParam$. Variational calibration therefore minimizes the expected Tikhonov functional minus the entropy. The expected Tikhonov term favors concentration in regions where $J$ is small, while the entropy term favors dispersion. Although the exact posterior minimizes the negative ELBO over all distributions, a restricted variational family and its optimization may yield an approximation concentrated near a single mode, particularly when modes are separated by regions of low probability mass.

\section{Prior R\'enyi regularization for variational inference}
\label{sec:renyi_regularization}

\subsection{Regularized objective}

The regularization aims to discourage the variational distribution from assigning too little mass to regions with appreciable prior probability. We employ the R\'enyi divergence
\begin{equation}\label{eq:renyi_definition}
\RenyiCall{\RenyiParam}{p}{q}
=\frac{1}{\RenyiParam-1}\log\int p(\IntVar)^{\RenyiParam}q(\IntVar)^{1-\RenyiParam}\,\dd\IntVar,
\qquad \RenyiParam>0,\quad\RenyiParam\ne1,
\end{equation}
which gives the regularized objective
\begin{equation}\label{eq:regularized_nelbo}
\RegNELBOCall{\VarParam}{\Obsn}{\Ctxn}
:=\NELBOof{\VarParam}{\Obsn}{\Ctxn}+\RegWeight\,\RegTerm{\VarParam},
\qquad \RegTerm{\VarParam}:=\RenyiCall{\RenyiParam}{p_{\rvParam}}{\qvar}.
\end{equation}
The direction from the prior to the approximation is chosen to discourage low variational density in regions carrying prior mass. For $\RenyiParam>1$, zero variational density on a set of positive prior probability gives an infinite divergence. The estimation of the regularization term is described in \cref{app:renyi_estimation}.

\subsection{Adaptive weight}

We balance the gradient magnitudes using a gradient-normalization strategy motivated by \cite{chen2018gradnorm}. At training step $t$, let $g_t^{(E)}=\|\nabla_\VarParam\NELBO\|_2$ and $g_t^{(R)}=\|\nabla_\VarParam\RegTerm{\VarParam}\|_2$. Their exponentially smoothed values and the weight update are
\begin{align}\label{eq:adaptive_regularization}
\bar g_t^{(a)}&=\EMAParam\bar g_{t-1}^{(a)}+(1-\EMAParam)g_t^{(a)},\qquad a\in\{E,R\},\nonumber\\
\RegWeight_{t+1}&=\GradRatio\frac{\bar g_t^{(E)}}{\bar g_t^{(R)}+\Stab},
\end{align}
with smoothing coefficient $\EMAParam\in[0,1)$, target gradient ratio $\GradRatio>0$, and numerical stabilizer $\Stab>0$. The updated weight is used at the next training step and is treated as fixed when computing the gradient of the regularized objective with respect to $\VarParam$.

\section{Numerical experiments}\label{sec:numerical_experiments}

\subsection{Synthetic benchmark}\label{sec:synthetic_benchmark}

We define the numerical model with parameter vector
\[
\Param=(\parami{0},\parami{1})\in\ParamSpace\subset\mathbb{R}^2,
\]
context variable
\[
\Ctx=(\ctxi{0},\ctxi{1})\in\CtxSpace\subset \Rdim{2},
\]
and observation space $\ObsSpace \subset \Rdim{10}$. The forward model is defined by
\begin{align}
\ModelObs : &\ParamSpace\times\CtxSpace\to\ObsSpace\\ \nonumber
            &\Param, \Ctx \mapsto \left(\left(\ModelObsCall{\Param}{\Ctx}\right)_{1}, \dots, \left(\ModelObsCall{\Param}{\Ctx}\right)_{\Obsdim}\right).
\end{align}
Each component of the forward model is defined separately. For $1 \leq i \leq 10$,
\begin{align}
\left(\ModelObsCall{\Param}{\Ctx}\right)_{i}
:=
\ctxi{1}\,\left( 1+\tfrac12\sin\!\left(\frac{i}{2}\right) \right)\,\parami{0}^2
-
\left( 1+\tfrac12\cos\!\left(\frac{i}{2}\right)\right)\parami{1}
+
\ctxi{0}.
\end{align}

In this experiment, calibration is performed using a synthetic observation at a single context $\CtxSingle$, given by $\ObsSingle=\ModelObsCall{\ParamTrue}{\CtxSingle}+\ObsNoise$, with diagonal Gaussian noise covariance $5\times10^{-2}\IdMat{10}$. By choosing $\CtxSingle$ and $\ParamTrue$, we construct two calibration cases. Case A produces a curved, V-shaped bimodal posterior with a substantial connecting region, while Case B produces modes separated by a region of very low probability density. The two configurations are summarized in \cref{tab:benchmark_cases}.
\begin{table}[!htbp]\centering
\caption{Twin-experiment configurations. Both cases use zero prior mean and observation covariance $5\times10^{-2}\IdMat{10}$. The prior covariance is denoted by $\CovPrior$.}\label{tab:benchmark_cases}
\begin{tabular}{lcc}\toprule
Case & True parameter $\ParamTrue$ & Context $\CtxSingle$\\\midrule
A & $(0.075,1)$ & $(1,1)$\\
B & $(0.4,1)$ & $(1,0.5)$\\\bottomrule
\end{tabular}\end{table}

\subsection{References and approximation diagnostics}\label{sec:baselines}

The reference posterior is obtained by evaluating the unnormalized density on a Cartesian grid and normalizing it by trapezoidal quadrature. Hamiltonian Monte Carlo provides an additional sampling-based comparison, with its density estimated by Gaussian kernel density estimation. The grid normalization procedure is described in \cref{app:reference_posteriors}.

For the reference posterior $\pi$ and approximation $\qvar$, we compute the forward and reverse Kullback--Leibler divergences and total variation distance:
\begin{align}\label{eq:diagnostics}
\DKL{\pi}{\qvar}&=\int\pi\log(\pi/\qvar),\qquad
\DKL{\qvar}{\pi}=\int\qvar\log(\qvar/\pi),\nonumber\\
\DTV{\pi}{\qvar}&=\tfrac12\int|\pi-\qvar|.
\end{align}
These quantities are computed using densities normalized on the same grid domain. When density evaluation at arbitrary points is intractable, the approximation is evaluated through kernel density estimation.

To quantify the fraction of runs exhibiting near collapse in Case B, we use the fact that the two modes lie on opposite sides of the first calibration coordinate. For run $i$, posterior samples are used to estimate the mass $\widehat p_{i,+}$ on the positive side of this coordinate and $\widehat p_{i,-}=1-\widehat p_{i,+}$ on the negative side. The minority-mode mass and near-collapse fraction are
\begin{equation}\label{eq:near_collapse}
\widehat m_i=\min(\widehat p_{i,+},\widehat p_{i,-}),\qquad
\widehat C_\epsilon=\frac1R\sum_{i=1}^{R}\mathbf1\{\widehat m_i<\epsilon\},\quad\epsilon=0.05.
\end{equation}
We also compute the squared energy distance \cite{szekely2013energy},
\begin{equation}\label{eq:energy_distance}
\mathcal E^2(P_X,P_Y)=2\E\|X-Y\|_2-\E\|X-X'\|_2-\E\|Y-Y'\|_2,
\end{equation}
where $X,X'$ are independent samples from $P_X$, and $Y,Y'$ are independent samples from $P_Y$, with all four random variables mutually independent.

\subsection{Numerical results}\label{sec:numerical_results}

\begin{figure}[!htbp]
  \centering
\includegraphics[width=370.39781000pt,height=165.49270000pt]{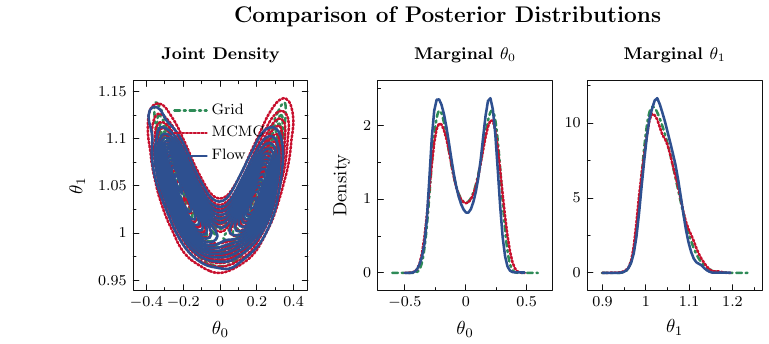}
  \caption{Case~A : Posterior with connected modes. Green: grid reference; red: Gaussian KDE of MCMC samples; blue: flow approximation. Left: joint-density contours. Middle and right: marginal densities.}
  \label{fig:posterior_connected}
\end{figure}

In \cref{fig:posterior_connected}, the variational approximation represents both modes and follows the grid and MCMC references. The final discrepancy measures are presented in \cref{tab:unregularized_metrics}.

\begin{figure}[!htbp]
    \centering
    \includegraphics[width=370.38715000pt,height=219.70218000pt]{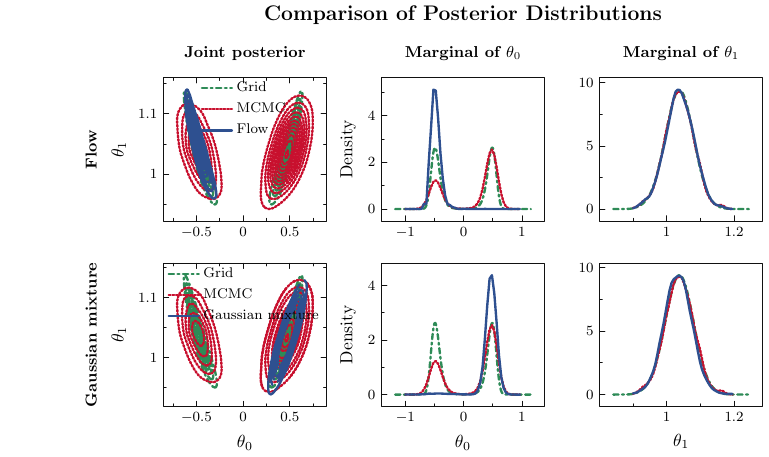}

    \caption{
    Case~B: posterior with well-separated modes.
    The first row shows the flow approximation and the second row
    the Gaussian mixture approximation.
    Green: grid reference; red: Gaussian KDE of MCMC samples;
    blue: variational approximation.
    The left column shows the joint-density contours, while the
    middle and right columns show the marginal densities.
    }

    \label{fig:posterior_disconnected_comparison}
\end{figure}

The same flow architecture frequently converges to one mode in Case B. The near-collapse fraction is $0.875$ in \cref{tab:unregularized_metrics}. The posterior are presented in \cref{fig:posterior_disconnected_comparison}. The Gaussian-mixture approximation shown in the same figure also converges to a single mode in the experiments with eight random seeds.

\begin{table}[!htbp]\centering\small
\caption{Unregularized posterior approximation. Divergences and distances are presented as mean $\pm$ standard deviation; the near-collapse fraction uses threshold $\epsilon=0.05$. }\label{tab:unregularized_metrics}
\begin{tabular}{lccc}\toprule
Metric & Case A flow & Case B flow & Case B mixture\\\midrule
Forward KL & $0.308\pm0.105$ & $11.197\pm4.940$ & $13.513\pm0.957$\\
Reverse KL & $0.474\pm0.079$ & $1.070\pm0.701$ & $15.592\pm0.277$\\
Total variation & $0.273\pm0.040$ & $0.512\pm0.010$ & $0.868\pm0.004$\\
Near-collapse fraction & -- & $0.875$ & $1.000$\\\bottomrule
\end{tabular}\end{table}

For the regularized approximation, we consider R\'enyi orders $\{0.1,0.4,0.8,1.2,1.6,2.0\}$ and gradient ratios $\{0.1,0.2,0.3\}$, with $\EMAParam=0.8$. The combination $\RenyiParam=0.8$ and $\GradRatio=0.2$ gives favorable results in terms of near-collapse fraction, minority-mode mass, and squared energy distance. Extended results for all parameter values are presented in \cref{app:regularization_study}.

For the matched-flow comparison, the unregularized reference and regularized approximation use the same architecture, which differs from the initial unregularized Case B flow. With $\RenyiParam=0.8$ and $\GradRatio=0.2$, regularization reduces the near-collapse fraction from $0.50$ to $0.250$. Forward Kullback--Leibler divergence decreases from $4.912$ to $3.126$ and squared energy distance from $0.259$ to $0.227$. Reverse divergence increases from $2.401$ to $3.358$, and total variation from $0.554$ to $0.617$. The means and standard deviations are given in \cref{tab:disconnected_reference_vs_renyi_metrics}.

\begin{table}[!htbp]
\centering
\begin{adjustbox}{max width=\linewidth}
\begin{tabular}{lcc}
\hline
\textbf{Metric}
& \textbf{Reference}
& \textbf{R\'enyi} \\
\hline

Forward KL
& \num{4.912} $\pm$ \num{4.867}
& \textbf{\num{3.126}} $\pm$ \textbf{\num{2.723}} \\

Reverse KL
& \textbf{\num{2.401}} $\pm$ \textbf{\num{1.850}}
& \num{3.358} $\pm$ \num{1.970} \\

Total Variation Distance
& \textbf{\num{0.554}} $\pm$ \textbf{\num{0.055}}
& \num{0.617} $\pm$ \num{0.075} \\

Squared Energy Distance
& \num{0.259} $\pm$ \num{0.174}
& \textbf{\num{0.227}} $\pm$ \textbf{\num{0.145}} \\

Near-collapse Fraction ($\epsilon = \num{5}\%$)
& \num{0.50}
& \textbf{\num{0.250}} \\

\hline
\end{tabular}
\end{adjustbox}
\caption{Case B: matched-reference and R\'enyi-regularized approximations. Values are mean $\pm$ standard deviation, except the near-collapse fraction. Lower values are preferred for all metrics.}
\label{tab:disconnected_reference_vs_renyi_metrics}
\end{table}

\begin{figure}[!htbp]
    \centering
            \includegraphics[width=370.38869000pt,height=273.59383000pt]{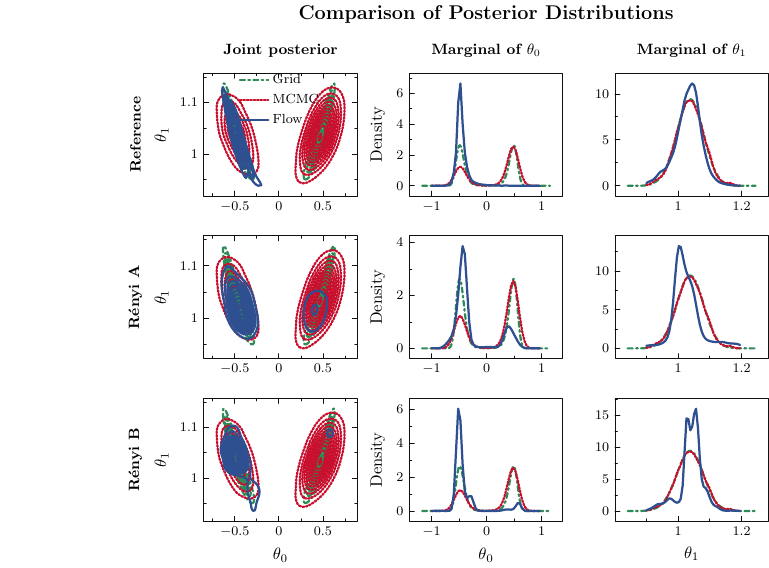}
    \caption{Case B: Comparison of the posterior distributions obtained
    with the reference training configuration and two
    R\'enyi-regularized runs. The rows correspond to the reference
    configuration, R\'enyi A, and R\'enyi B, with the latter two
    corresponding to different random seeds. The columns show the joint
    posterior distribution and the two marginal posterior distributions.}
\label{fig:posterior_comparison_renyi_regularization}
\end{figure}

The two regularized realizations in \cref{fig:posterior_comparison_renyi_regularization} illustrate the remaining variability: R\'enyi A represents both modes, whereas R\'enyi B nearly collapses. Regularization reduces near collapse but does not prevent it. Details of the training configurations and flow architectures are provided in \cref{app:settings}.

\FloatBarrier

\section{Discussion and conclusion}

The negative ELBO combines the expected data misfit and prior deviation with a negative entropy term. In the symmetric benchmark, either mode explains the observation and is equally favored by the prior, so a distribution concentrated near one mode can have a small expected Tikhonov term. Entropy maximization alone does not enforce multimodality, since a unimodal Gaussian maximizes differential entropy at fixed covariance \cite{cover2005differential}. This is consistent with the observation that flow-based variational inference approximates the connected bimodal posterior but frequently misses one of the well-separated modes in Case B.

The flow and mixture results are consistent with the mode-seeking behavior of the reverse divergence, although they remain dependent on the architecture, initialization, and optimization settings. Adding the prior-referenced R\'enyi regularization reduces near collapse in the matched-flow comparison, but also changes the variational target and can spread probability into regions less favored by the data. In this comparison, forward divergence and energy distance decrease, whereas reverse divergence and total variation increase.

Overall, R\'enyi regularization can improve mode coverage, but its benefit must be assessed against the bias introduced by modifying the Bayesian objective and the variability of the optimization.

\FloatBarrier
\appendix
\section{Reference posteriors}\label[appendix]{app:reference_posteriors}

Let $\widetilde\pi_{jk}$ denote the unnormalized posterior evaluated on a Cartesian grid:
\begin{equation}
\widetilde\pi_{jk}
=
p_{\rvParam}(\parami{0}^j,\parami{1}^k)
\prod_{i=1}^{n}
p_{\rvObsi{i}\mid\rvParam,\rvCtxi{i}}
\!\left(
\Obsi{i}\mid(\parami{0}^j,\parami{1}^k),\Ctxi{i}
\right).
\end{equation}
Let $w_{jk}$ be the trapezoidal quadrature weights. The reference posterior and the evaluated approximation are normalized on the same grid domain:
\begin{equation}\label{eq:grid_normalization}
\pi_{jk}
=
\frac{\widetilde\pi_{jk}}{\sum_{j,k}w_{jk}\widetilde\pi_{jk}},
\qquad
q_{jk}
=
\frac{\qvar(\parami{0}^j,\parami{1}^k)}
{\sum_{j,k}w_{jk}\qvar(\parami{0}^j,\parami{1}^k)}.
\end{equation}
The integrals in \cref{eq:diagnostics} are approximated by the corresponding weighted sums. Gaussian kernel density estimation is used when density evaluation at arbitrary points is intractable.

\section{Experimental settings}
\label[appendix]{app:settings}

\begin{table}[!htbp]
\centering
\footnotesize
\setlength{\tabcolsep}{4pt}
\renewcommand{\arraystretch}{0.95}
\begin{adjustbox}{max width=\linewidth}
\begin{tabularx}{\textwidth}{@{}p{0.28\textwidth}Xp{0.35\textwidth}@{}}
\toprule
Configuration & Component & Value \\
\midrule

\textbf{Common flow settings}
& Number of seeds & \num{8} \\
& Learning rate & \num{1e-3} \\
& Number of training steps & \num{5000} \\
& Number of particles & \num{10} \\
& Flow hidden dimensions & $(\num{8},\num{8})$ \\
& Number of spline bins & \num{8} \\
& Spline bound & \num{5.0} \\
& Repetitions per transform block & \num{3} \\

\midrule
\textbf{Case A: connected modes}
& Architecture
& Spline coupling $\rightarrow$ spline autoregressive
  $\rightarrow$ neural autoregressive \\
& Neural autoregressive width & \num{8} \\

\midrule
\textbf{Case B: well-separated modes}
& Unregularized architecture
& Same as Case A \\

\midrule
\textbf{Case B: matched-flow comparison}
& Reference architecture
& Affine coupling $\rightarrow$ spline coupling
  $\rightarrow$ affine coupling $\rightarrow$ spline autoregressive
  $\rightarrow$ affine coupling \\
& Regularized posterior family
& Same as reference \\
& R\'enyi order $\rho$
& $\{\num{0.1},\num{0.4},\num{0.8},
     \num{1.2},\num{1.6},\num{2.0}\}$ \\
& ELBO-to-regularization gradient ratio $r$
& $\{\num{0.1},\num{0.2},\num{0.3}\}$ \\
& EMA coefficient $\beta$ & \num{0.8} \\

\midrule
\textbf{Gaussian mixture}
& Initial means
& $\left(
    (\num{-2},\num{-2}),
    (\num{2},\num{4})
  \right)$ \\
& Initial scale & \num{0.5} \\
& Learning rate & \num{1e-2} \\

\bottomrule
\end{tabularx}
\end{adjustbox}
\caption{Hyperparameters for the normalizing-flow and Gaussian-mixture
approximations. Common flow settings apply to the flow configurations.
The Case B matched-flow comparison uses the same architecture for
the unregularized reference and the R\'enyi-regularized approximation.}
\label{tab:hyperparams}
\end{table}

\FloatBarrier

\section{Monte Carlo estimation of the R\'enyi divergence}
\label{app:renyi_estimation}

For independent samples $\IntVar_i\sim p_{\rvParam}$, set
$s_i=(\RenyiParam-1)[\log p_{\rvParam}(\IntVar_i)-\log\qvar(\IntVar_i)]$. The prior-expectation form of the divergence and its log-sum-exp estimator \cite{li_renyi_2016} are
\begin{align}\label{eq:renyi_mc}
\RenyiCall{\RenyiParam}{p_{\rvParam}}{\qvar}
&=\frac{1}{\RenyiParam-1}\log\E_{p_{\rvParam}}\!\left[\left(\frac{p_{\rvParam}}{\qvar}\right)^{\RenyiParam-1}\right],\nonumber\\
\widehat{\mathcal D}_{\mathrm{Renyi}}^{\RenyiParam}
&=\frac{\operatorname{logsumexp}_{i=1}^{N}(s_i)-\log N}{\RenyiParam-1}.
\end{align}
The estimator is generally biased because of the logarithm but is consistent when the expectation is finite. Since the prior samples are independent of $\VarParam$, differentiation yields
\begin{equation}\label{eq:renyi_grad}
\nabla_\VarParam\widehat{\mathcal D}_{\mathrm{Renyi}}^{\RenyiParam}
=-\sum_{i=1}^{N}\frac{\exp(s_i)}{\sum_j\exp(s_j)}\nabla_\VarParam\log\qvar(\IntVar_i).
\end{equation}

\section{Regularization parameter study}
\label[appendix]{app:regularization_study}

\begin{figure}[!htbp]
    \centering
\includegraphics[width=370.39832000pt,height=163.58133000pt]{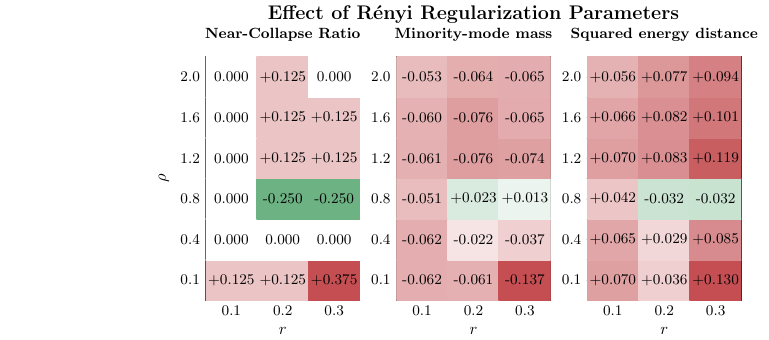}
    \caption{Changes in the evaluation metrics for the disconnected
    posterior across R\'{e}nyi orders $\RenyiParam$ and gradient
    ratios $\GradRatio$. Cell labels report the changes relative
    to the reference configuration; green indicates improvement,
    whereas red indicates deterioration.}
    \label{fig:renyi_parameter_heatmaps}
\end{figure}

\FloatBarrier
\section*{Acknowledgments}

This work was supported by the French Ministry of
Higher Education and Research through the CIFRE
(Conventions Industrielles de Formation par la Recherche)
program, administered by the Association Nationale de
la Recherche et de la Technologie (ANRT), under
CIFRE agreement No. 2022/1801, in collaboration
with Bull.

This project was provided with HPC/AI computing and
storage resources by GENCI at IDRIS thanks to grant
AD010117241 on the Jean Zay supercomputer's CSL,
A100, and H100 partitions.
\bibliographystyle{siamplain}
\bibliography{references}
\end{document}